\documentclass[letterpaper, 10 pt, conference]{ieeeconf}  

\IEEEoverridecommandlockouts                     
\usepackage{graphicx}       
\usepackage{amsmath}        
\usepackage{amssymb}      
\usepackage{mathptmx}       
\usepackage{times}          
\usepackage{booktabs}      
\usepackage{pifont}         
\usepackage{tabularx}   
\usepackage{threeparttable}
\usepackage{balance}
\usepackage[caption=false,font=footnotesize]{subfig}
\newcolumntype{Y}{>{\raggedright\arraybackslash}X}

\usepackage[table]{xcolor} 
\usepackage{cite}    
\usepackage[hidelinks]{hyperref}
\usepackage{makecell}
\title{\LARGE \bf
When to Personalize Household Object Search: A Rigidity-Gated Hybrid Policy}

\author{Xianyao Li$^{1}$, Yuhai Wang$^{2}$, Hu Xiao$^{2}$, Kaleb Smith$^{3}$, Gilbert Yang Ye$^{2}$ and Eric Jing Du$^{1}$%
\thanks{$^{1}$X. Li and E. J. Du are with the Department of Civil and Coastal Engineering, University of Florida, Gainesville, FL 32611, USA. {\tt\small xianyao.li@ufl.edu}, {\tt\small eric.du@essie.ufl.edu}. $^{2}$Y. Wang, H. Xiao, and G. Y. Ye are with the Department of Civil and Environmental Engineering, Northeastern University, Boston, MA 02115, USA. $^{3}$K. Smith is with NVIDIA, {\tt\small kasmith@nvidia.com}. Project resources: {\tt\small \url{https://github.com/XianyaoLi/PerSim}}.}
}

\begin{document}

\maketitle

\thispagestyle{empty}
\pagestyle{empty}

\begin{abstract}
Service robots searching for household objects rely on spatial priors to reduce search cost, yet object locations can vary with resident traits. Collecting longitudinal, trait-specific in-home trajectories is invasive and hard to scale. We study \emph{when} personalization helps and propose \emph{PerSim}, a \emph{rigidity-gated hybrid policy} that combines a \emph{trait-conditioned} prior with a \emph{population-frequency baseline}, personalizing only when placement behavior is variable.
To scale resident-conditioned dynamics, we employ a human-calibrated simulation pipeline to generate and validate object-placement transitions in diverse home layouts, and train a predictor that injects continuous Big Five vectors to output room-level priors and within-room co-occurrence cues.
In a unified human study ($N=200$), dual-layer validation shows that (i) synthetic transitions are judged behaviorally plausible (mean $3.85/5$, $p<10^{-6}$), and (ii) in a blinded A/B comparison, personalization is favored primarily for low-rigidity objects ($p\!=\!0.005$), while the population-frequency baseline remains strong for universally placed items---yielding a decision rule for when to personalize. In an offline objective test, we observe a small but significant improvement on unseen continuous trait vectors over nearest discrete configuration matching ($p\!=\!0.035$), supporting interpolation in five-dimensional trait space. Finally, in a home digital twin we show that PerSim reduces \emph{expected search cost} by combining room visitation effort with within-room cue checking, demonstrating end-to-end gains beyond isolated prediction metrics.
\end{abstract}

\section{Introduction}
\label{sec:introduction}

In everyday homes, small items go missing constantly: a mug gets left in the bedroom, a phone slips between couch cushions, or keys end up on an unexpected surface. Home service robots must therefore search intelligently---when asked to fetch a mug, phone, or keys, a robot should prioritize likely locations rather than exhaustively scanning every room~\cite{zhang2019efficient, pangercic2012semantic, chaplot2020object}. Spatial priors make this possible. Yet household object locations are not determined by environment semantics alone (e.g., mugs near kitchens); they are also shaped by \emph{resident traits}. Some residents consistently return items to fixed storage, while others tolerate functional clutter and frequent relocation. This creates a practical dilemma for robotics: \emph{personalization can help---but not always}. If a robot personalizes aggressively when behavior is stable and universally shared, it risks overfitting noise; if it never personalizes when behavior is variable and resident-dependent, it wastes an opportunity to reduce search cost.

\begin{figure}[t]
    \centering
    \includegraphics[width=\columnwidth, trim=0 0 0 0, clip]{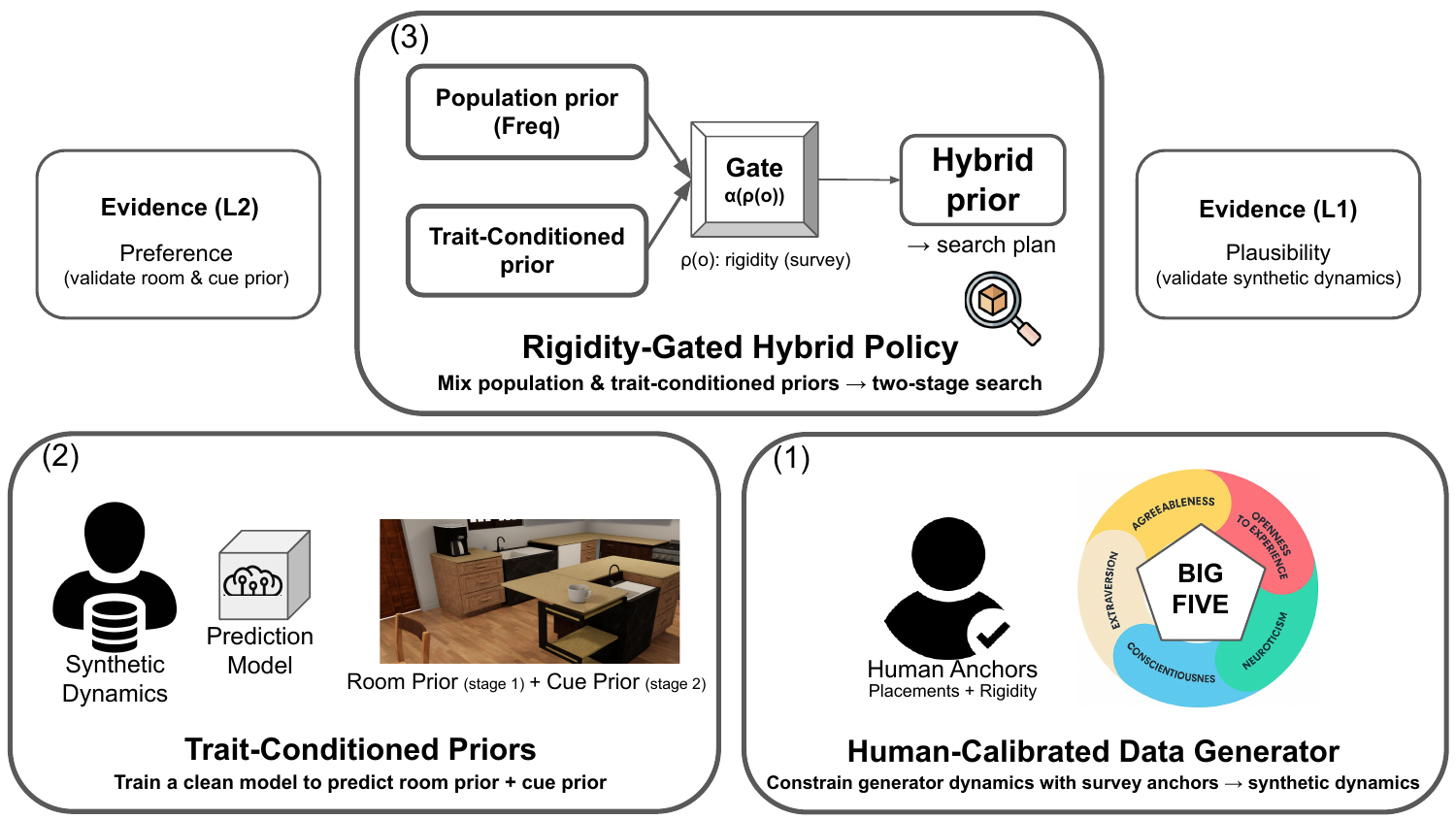}
    \caption{\textbf{PerSim as a hypothesis-driven framework for rigidity-gated personalization.}
(1) Human anchors provide resident profiles and object-level placement/rigidity signals to calibrate a constrained generative model, producing behaviorally plausible synthetic dynamics (validated by L1).
(2) A clean predictor learns \emph{trait-conditioned} \emph{room priors} and \emph{cue priors} for two-stage search (stage~1: room ranking; stage~2: within-room cueing), whose outputs are validated by a blinded preference study (L2).
(3) A rigidity-gated hybrid policy mixes population (frequency) and trait-conditioned priors via $\alpha(\rho(o))$ to decide when personalization is beneficial, yielding a robot-facing search plan.
Scene snapshots are from BEHAVIOR-1K~\cite{li2024behavior}.}
\label{fig:pipeline}
\end{figure}

This paper asks a focused question: \textbf{when should a robot personalize household object search?} We argue that the answer hinges on \emph{placement rigidity}: some objects exhibit consistent, universally shared placement patterns, while others move fluidly across rooms and contexts. This motivates PerSim, a \emph{rigidity-gated hybrid policy} that selectively personalizes only when trait-conditioned information is likely to pay off, and otherwise defaults to a population-frequency baseline.
To study this decision in a setting that reflects everyday ``lost-and-found'' behavior, we evaluate on a controlled query set of 15 common, easily misplaced household items that users often need to actively search for, chosen to span the rigidity spectrum from anchor-like to trait-sensitive objects.
Fig.~\ref{fig:pipeline} overviews the PerSim framework.

Progress on this question is constrained by a data bottleneck. Longitudinal in-home collection of resident-specific placements is invasive and hard to scale, while many embodied datasets emphasize static layouts or scripted activities and under-represent repeated relocation across days and contexts~\cite{li2024behavior}. As a result, models often learn population-level regularities but lack the signals needed to learn \emph{trait-conditioned} placement dynamics and, crucially, to quantify when such conditioning is beneficial.

Large language models (LLMs) offer a tempting way to synthesize household behaviors at scale~\cite{park2023generative, ahn2022saycan}. However, for downstream robotic prediction, a core risk is distribution mismatch: unconstrained generations may reflect an LLM's internal commonsense prior rather than the empirical distribution of how residents actually place objects, producing biased priors that can harm search.

We present \textit{PerSim}, a human-calibrated framework for learning \emph{trait-conditioned} household object search priors from validated placement transitions in an OmniGibson-based~\cite{li2024behavior} home digital twin (Fig.~\ref{fig:pipeline}). PerSim operationalizes resident traits using continuous Big Five vectors~\cite{john1999bigfive}, enabling interpolation across individuals rather than assigning discrete resident types. PerSim uses human anchor records to calibrate a constrained generative model that produces behaviorally plausible object-placement transitions; these synthetic transitions are then used to train a clean \emph{trait-conditioned} predictor that outputs (i) room-level priors and (ii) within-room cue priors (co-occurrence cues) to support a two-stage search strategy.

Our central claim is not that personalization helps everywhere, but that its utility is \emph{heterogeneous}. In a unified human study protocol, we obtain dual-layer evidence: first, generated transitions are judged behaviorally plausible; second, in blinded A/B comparisons, personalization is favored primarily for \emph{low-rigidity} objects, while the population-frequency baseline remains strong for universally placed items. This \emph{rigidity-modulated gradient} yields a practical decision rule: \textbf{use a rigidity gate to mix a population-frequency baseline with trait-conditioned priors}. Finally, in a home digital twin, we show that this rigidity-gated hybrid policy reduces expected search cost by jointly accounting for room visitation effort and within-room cue checking, demonstrating end-to-end gains beyond isolated prediction metrics.

Our contributions are:
\begin{itemize}
    \item \textbf{Trait-conditioned priors for household object search:} a predictor that produces room-level priors and within-room cue priors to support two-stage robot search.
    \item \textbf{Rigidity-gated hybrid policy (when to personalize):} dual-layer human evidence that trait conditioning is most beneficial for low-rigidity objects, motivating a policy that combines trait-conditioned priors with a population-frequency baseline via a rigidity gate.
    \item \textbf{Human-calibrated synthetic dynamics:} a scalable pipeline that aligns LLM generation with human anchors to produce validated placement transitions for learning trait-conditioned dynamics.
    \item \textbf{Interpolation in continuous trait space:} evidence that conditioning on continuous Big Five trait vectors improves generalization to unseen traits compared with nearest discrete matching.
\end{itemize}

\section{Related Work}
\label{sec:related_work}

\paragraph{Semantic priors for household object search}
Service robots often search for household objects by exploiting semantic and spatial regularities: object--room affinities (e.g., mugs in kitchens), contextual cues, and co-occurrence structure that narrow down where to look first~\cite{pangercic2012semantic, zhang2019efficient, 2024seek, sarch2022tidee, kant2022housekeep}. These priors have enabled efficient search under population-level regularities, but they are typically \emph{not trait-conditioned}: they rarely account for stable individual differences in organization style or for multi-day relocation dynamics that determine where an object is \emph{currently} likely to be in a given home. Our work complements semantic priors by learning \emph{trait-conditioned} room and within-room cue priors, and by explicitly studying \emph{when} trait conditioning improves search via a rigidity-gated hybrid policy.

\paragraph{Personalization in human environments and trait-conditioned resident modeling}
Personalization for agents in human environments has been explored through routines, preferences, and user modeling, where stable individual differences shape how people organize personal spaces~\cite{gosling2002room, john1999bigfive, abdo2016organizing, wu2023tidybot}. In household robotics, such differences motivate \emph{trait-conditioned} search priors instead of one-size-fits-all semantics. A recurring bottleneck, however, is supervision at the right granularity and timescale: collecting repeated, longitudinal in-home object placements is invasive and costly, and many embodied datasets emphasize static layouts or scripted activities, often lacking the resident-specific annotations and multi-day dynamics needed to study when personalization is beneficial~\cite{li2024behavior}. We address this gap by operationalizing resident traits with continuous Big Five vectors (OCEAN) and by focusing the objective on robot-centric decision making: a principled rule for when to personalize search.

\paragraph{LLM-assisted behavior synthesis and calibration for robotic priors}
Large language models (LLMs) have been used to produce plans, task decompositions, and high-level behaviors for embodied agents~\cite{ahn2022saycan, park2023generative, wang2023voyager, huang2023inner}. For learning robotic priors from synthetic data, a central risk is \emph{distribution mismatch}: unconstrained generations may reflect an LLM's internal commonsense rather than the empirical distribution of household routines, inducing biased priors that can harm downstream search. PerSim mitigates this risk by anchoring synthesis to human placement records, training a constrained generator (via supervised alignment), and validating outputs with a unified dual-layer protocol that checks both behavioral plausibility and downstream utility for search.

\paragraph{When to personalize: gating and hybridization}
A separate but closely related theme is that personalization is not uniformly helpful. Prior work in robotics and human-centered learning often blends generic and user-specific models or adapts selectively when evidence supports it. Our work makes this principle explicit for household object search by introducing a \emph{rigidity-gated hybrid policy} that mixes a \emph{population-frequency baseline} with \emph{trait-conditioned} priors. The key distinction is that our gate is grounded in object placement rigidity: it is designed to default to population regularities for universally placed objects and to invoke trait conditioning primarily for low-rigidity objects, matching the heterogeneous preference patterns observed in our human study.

\begin{figure*}[t]
    \centering
    \includegraphics[width=0.98\textwidth, trim=0 0 0 145, clip]{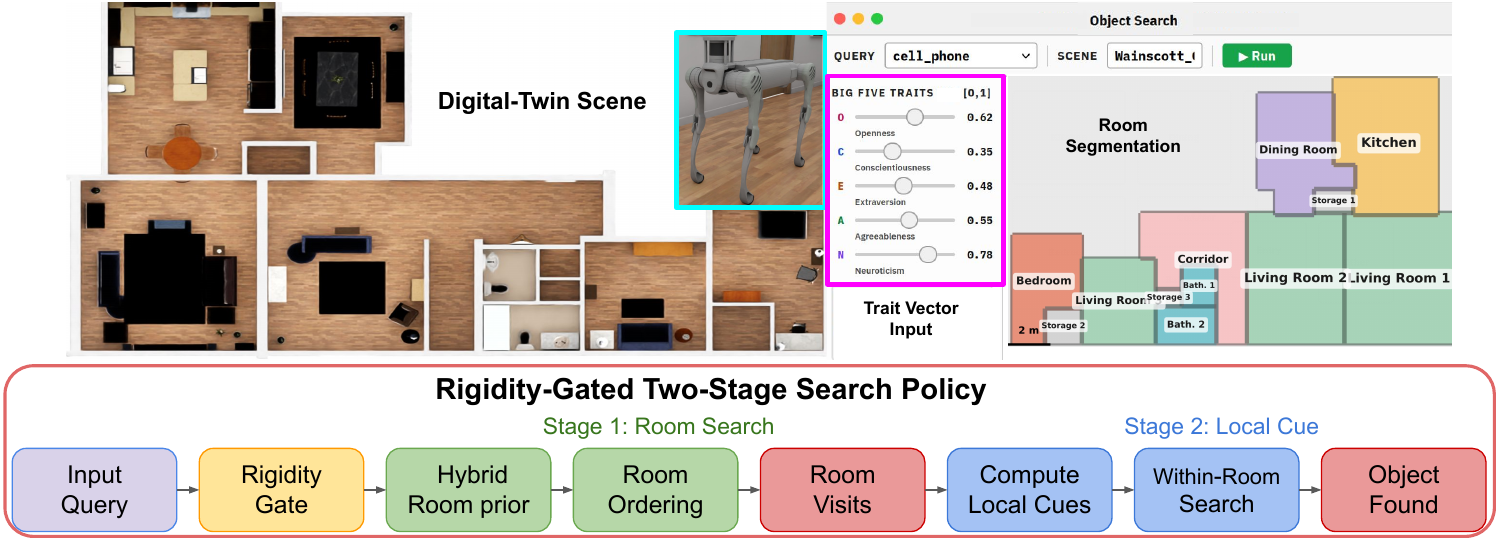}
    \caption{\textbf{Rigidity-gated two-stage digital-twin search policy.}
Given an object query and a trait vector, a rigidity gate determines how much to rely on a population room prior versus a trait-conditioned prior, producing a hybrid room ranking for Stage~1 search (lower ERV is better).
After entering the target room, Stage~2 performs local search by inspecting the top-$K$ predicted co-occurrence cues (reported as CP@5; higher is better).
Scenes are from BEHAVIOR-1K~\cite{li2024behavior}.}
\label{fig:hybrid_policy}
\end{figure*}

\section{PerSim: Human-Calibrated Trait-Space Simulation and Trait-Conditioned Priors}
\label{sec:method}

PerSim is designed around a testable claim: \emph{trait conditioning improves robot search primarily when object placement is behaviorally variable (low rigidity), and provides limited benefit when placement is universal (high rigidity).}
Accordingly, PerSim (i) learns \emph{trait-conditioned} search priors from scalable, human-calibrated multi-day placement dynamics, and (ii) deploys them through a \emph{rigidity-gated hybrid policy} that defaults to a strong \emph{population-frequency baseline} when personalization is unlikely to help.

\subsection{Robot-Facing Task and Predicted Priors}
\label{subsec:problem_setup}

PerSim targets \emph{robotic household object search}. Let $\mathbf{t}\in\mathbb{R}^5$ denote a resident's continuous Big Five trait vector (OCEAN).
Each household scene $s$ contains rooms $\mathcal{R}_s$ and a movable-object vocabulary $\mathcal{O}$ (in our selected BEHAVIOR-1K scenes, $|\mathcal{O}|=197$).
For a query object $o$, PerSim predicts:
(i) a \textbf{room prior} $\hat{p}(r\mid s,o,\mathbf{t})$ over $r\in\mathcal{R}_s$, and
(ii) a \textbf{within-room cue prior} $\hat{\mathbf{y}}(o)$ over movable objects, representing likely co-occurring cues once a room is entered.
These priors support a two-stage search strategy: rank rooms first, then inspect top-ranked cues within the selected room.

\paragraph{Rigidity and ``when to personalize''}
Each query object is associated with a rigidity score $\rho(o)$ (Sec.~\ref{subsec:human_anchors}, Table~\ref{tab:object_rigidity}).
Rigidity operationalizes behavioral variability: high-rigidity objects tend to exhibit stable, widely shared placements; low-rigidity objects exhibit resident- and context-dependent relocation.
PerSim uses $\rho(o)$ to decide \emph{when} trait conditioning is likely to pay off, motivating a deployable hybrid policy rather than uniform personalization.

\subsection{Policy Instantiation and Robot-Centric Metrics}
\label{subsec:dt_policy}

To connect predicted priors to robot behavior, we instantiate a simple two-stage search policy in a home digital twin (OmniGibson), illustrated in Fig.~\ref{fig:hybrid_policy}.
Given a query $(o,s,\mathbf{t})$, a method provides (i) a room distribution $\hat{p}(r\mid s,o,\mathbf{t})$ and (ii) cue scores over movable objects.

\paragraph{Stage 1: room ordering and ERV}
Rooms are ranked by descending $\hat{p}(r\mid s,o,\mathbf{t})$ and searched in this order until the simulator ground-truth room $r^\star$ is reached.
We quantify efficiency by \textbf{Expected Rooms Visited (ERV)}:
\begin{equation}
\mathrm{ERV}=\mathbb{E}\big[\mathrm{rank}(r^\star)\big],
\end{equation}
where $\mathrm{rank}(r^\star)$ is the 1-indexed position of $r^\star$ in the ranked list (lower is better).

\paragraph{Stage 2: within-room cueing and CP@5}
Conditioned on $r^\star$, we rank candidate cue objects by predicted co-occurrence scores and prioritize checking the top-$K$ cues ($K=5$).
Ground-truth neighbors are defined by simulator-derived same-room proximity within radius $\delta$ (Sec.~\ref{subsec:relation_cooccurrence}).
We quantify cue usefulness via \textbf{Cue Precision@5 (CP@5)}:
\begin{equation}
\mathrm{CP@5}=\frac{\left|\,\mathrm{top}\text{-}5\ \cap\ N_{\delta}(o)\,\right|}{5}.
\end{equation}
We condition Stage~2 on $r^\star$ to isolate local cue-ranking quality; end-to-end search composes Stage~1 and Stage~2 sequentially.
We also report \textbf{Expected Search Cost (ESC)} as a unified proxy combining ERV with wasted within-room cue checks derived from CP@5 (Sec.~\ref{sec:dt_search}).

\paragraph{Rigidity-gated hybrid policy}
PerSim combines a strong \emph{population-frequency baseline} with a \emph{trait-conditioned} prior through a rigidity gate.
Let $\hat{p}_{p}$ denote the population (trait-agnostic) room prior, and $\hat{p}_{t}$ the trait-conditioned room prior.
We define the hybrid prior
\begin{equation}
\begin{split}
\hat{p}_{\mathrm{HYB}}(r\mid s,o,\mathbf{t})=\;&\alpha(\rho(o))\,\hat{p}_{t}(r\mid s,o,\mathbf{t})\\
&+(1-\alpha(\rho(o)))\,\hat{p}_{p}(r\mid s,o),
\end{split}
\end{equation}
where the mixing weight $\alpha\in[0,1]$ is deterministically mapped from rigidity via a clipped linear schedule:
\begin{equation}
\label{eq:alpha}
\alpha(\rho)=\mathrm{clip}\!\left(\frac{\tau_H-\rho}{\tau_H-\tau_L},\,0,\,1\right).
\end{equation}
$\alpha(\rho)$ is monotone decreasing in rigidity, mapping low-rigidity objects to $\alpha\approx 1$ and anchors to $\alpha\approx 0$.
We set $(\tau_L,\tau_H)=(3.8,4.3)$ in this study (Table~\ref{tab:object_rigidity}); these thresholds sit at natural gaps in the survey-measured rigidity distribution, not tuned on outcomes. Since $\alpha(\rho)$ varies smoothly (Eq.~\ref{eq:alpha}), small threshold shifts perturb the gate only gradually rather than abruptly.
Deployments can recalibrate thresholds using light feedback or held-out data.

\subsection{Human Anchors: Traits and Placement Supervision}
\label{subsec:human_anchors}

PerSim is grounded in a unified human study that provides (i) resident trait measurements and (ii) placement anchors used for calibration and evaluation.
We recruited $N=200$ adult participants via Prolific (\emph{Round~1}).

\paragraph{Traits}
Participants complete BFI-10~\cite{rammstedt2007bfi10}, yielding a continuous Big Five trait vector $\mathbf{t}\in\mathbb{R}^5$~\cite{john1999bigfive}.
We use the term \emph{trait-conditioned} to denote any model component that conditions on $\mathbf{t}$.

\begin{table}[t]
\centering
\small
\caption{15 query objects grouped by placement rigidity.}
\label{tab:object_rigidity}
\setlength{\tabcolsep}{4.0pt}
\renewcommand{\arraystretch}{0.92}
\begin{tabularx}{\columnwidth}{@{}lX@{}}
\toprule
\textbf{Type (Rigidity)} & \textbf{Objects} \\
\midrule
A\enspace(Anchor, $\rho \ge \tau_H\;(4.3)$) &
toothbrush, dish\_soap, pillow, blanket \\[2pt]
\multicolumn{1}{@{}p{4.2cm}}{C\enspace(Moderate, $\tau_L\;(3.8) < \rho < \tau_H\;(4.3)$)} &
mug, laptop, backpack, wallet, rag, charger, plate, key \\[2pt]
B\enspace(Sensitive, $\rho \le \tau_L\;(3.8)$) &
cell\_phone, water\_bottle, notebook \\
\bottomrule
\multicolumn{2}{@{}l}{\scriptsize Rigidity: 1--5 Likert (1=random, 5=fixed).}
\end{tabularx}
\end{table}

\paragraph{Placement anchors (15 query objects)}
For each of 15 everyday objects (Table~\ref{tab:object_rigidity}), participants report:
(i) a \textbf{room placement} distribution over 8 standardized room types,
(ii) a \textbf{nearby movable-object} multi-select set (co-occurrence cues), and
(iii) \textbf{rigidity} on a 1--5 Likert scale (1: random; 5: fixed).
This yields $200\times 15=3{,}000$ anchored object records.
We do not claim the 15 objects exhaust household categories; they form a representative controlled query set spanning the rigidity spectrum to test rigidity-modulated personalization under a single protocol.
Objects are selected to exist across typical residential layouts in the BEHAVIOR-1K scenes used for downstream simulation~\cite{li2024behavior}.

\paragraph{Quality control}
We apply automated checks (catch trials, straight-lining, implausible ratings, and anomalous completion times) and exclude low-quality submissions.
The same cohort is used for dual-layer validation (Sec.~\ref{subsec:dual_layer_protocol}) in a separate \emph{Round~2}, distinct from \emph{Round~1}.

\subsection{Human-Calibrated Synthetic Multi-day Placement Dynamics}
\label{subsec:llm_calibration}
Longitudinal, trait-conditioned placement dynamics are difficult to collect in real homes at scale.
We therefore synthesize multi-day object-placement transitions, but enforce \emph{calibration} to reduce distribution mismatch.
The goal is not ``creative'' behavior synthesis, but trajectories whose induced priors align with human judgments and support robot search.

\paragraph{Anchored, structured generation}
We train a constrained generator on the human anchors (Sec.~\ref{subsec:human_anchors}).
Specifically, we perform supervised fine-tuning (SFT) of Gemini~2.5~Flash on the \emph{Round~1} anchor data via Google Cloud Vertex AI to reduce distribution mismatch and stabilize structured outputs; \emph{Round~2} dual-layer validation uses separate stimuli (Sec.~\ref{subsec:dual_layer_protocol}).
Given a scene schema (room inventory and movable vocabulary), a query object, and a trait vector $\mathbf{t}$, the generator produces structured outputs for
(i) room placement and (ii) within-room co-occurrence cues, for an initial layout and a next-day update.
Structured outputs (room labels and cue sets) enable automatic validation and downstream relation extraction.

\paragraph{Trait-space coverage (orthogonal design)}
To cover the continuous five-dimensional trait space efficiently, we sample 16 representative trait configurations using an $L_{16}(2^5)$ Resolution-V orthogonal array~\cite{unal1990taguchi, kacker1991taguchi}.
Each trait takes two levels at population mean $\pm 1.0$ SD, supporting disentangled estimation of trait effects with limited sampling.
We generate day-by-day updates over $D=14$ days across a fixed pool of five BEHAVIOR-1K residential scenes~\cite{li2024behavior} to avoid confounding traits with environments.

\paragraph{Automatic validation and dataset summary}
We enforce valid room IDs, canonical object names, and complete structured fields required for simulation and relation extraction, retaining only validated transitions.
This yields 27{,}976 validated object-level actions across configurations, scenes, and days.

\subsection{Simulation Grounding and Co-occurrence Construction}
\label{subsec:relation_cooccurrence}

We instantiate each validated layout in OmniGibson and execute generated trajectories in simulation.
At each day, we extract object centroids and room membership from simulator state to compute metric distances and local neighborhoods.

\paragraph{Co-occurrence definition}
For query object $o$ at day $d$, we define the co-occurrence set as same-room movable objects within radius $\delta$:
\begin{equation}
c_d(o)=\{\,o'\in\mathcal{O}\setminus\{o\}\;:\; r_d(o')=r_d(o)\ \wedge\ \mathrm{dist}(o,o')\le \delta \,\}.
\end{equation}
We encode $c_d(o)$ as a multi-hot vector $\mathbf{y}_d(o)\in\{0,1\}^{|\mathcal{O}|}$.
We set $\delta=2.5$m based on the empirical same-room neighbor-distance distribution in our simulated dataset and use it consistently across scenes.

\paragraph{Room schema}
Room labels are normalized to 8 types to align survey supervision with simulation outputs, ensuring consistent room-prior prediction and evaluation.

\begin{figure*}[t]
  \centering
  \includegraphics[width=0.95\textwidth]{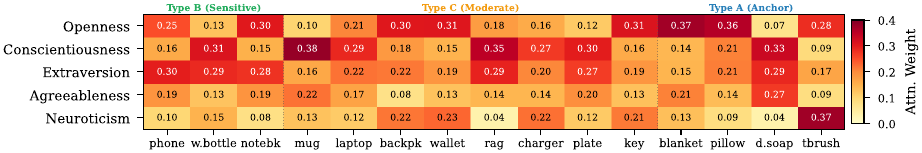}
  \caption{\textbf{Object-specific dependence on personality dimensions.}
  Cross-attention weights over Big Five traits (ordered as OCEAN) vary systematically across query objects,
  indicating that the trait-conditioned prior relies on different trait signals for different items.
  Objects are grouped by rigidity type (A/B/C) to align with the rigidity-gated policy.}
  \label{fig:trait_attention}
\end{figure*}

\begin{figure*}[t]
  \centering
  \includegraphics[width=0.95\textwidth]{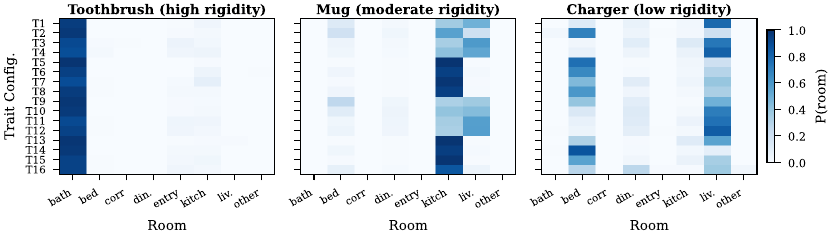}
  \caption{\textbf{Three regimes of trait-conditioned room priors.}
  Room distributions across 16 orthogonal trait configurations (T1--T16) illustrate rigidity-modulated behavior:
  \textit{toothbrush} is anchor-like (stable unimodal prior),
  \textit{mug} is moderate (dominant mode with limited shifts),
  and \textit{charger} is low-rigidity/trait-sensitive (multi-modal prior).
  This qualitative pattern motivates gating personalization by object rigidity.}
  \label{fig:room_priors_regimes}
\end{figure*}

\subsection{Trait-Conditioned Placement Predictor}
\label{subsec:predictor}

We train a clean trait-conditioned predictor on validated simulated actions to estimate robot-facing search priors.
Given an object category, lightweight temporal context, and a continuous trait vector $\mathbf{t}\in\mathbb{R}^5$, the model outputs
(i) a room distribution over $\mathcal{R}_s$ and
(ii) multi-label co-occurrence logits over $\mathcal{O}$.

\paragraph{Trait conditioning}
To inject $\mathbf{t}$ as a continuous control signal, we project each trait dimension $t_i$ to a token $\mathbf{z}_i$ and apply a cross-attention block~\cite{vaswani2017attention} where the context embedding queries trait tokens:
\begin{equation}
\mathbf{h}'=\mathrm{LN}\big(\mathbf{h}+\mathrm{Attn}(\mathbf{h},\mathbf{Z},\mathbf{Z})\big),\quad \mathbf{Z}=[\mathbf{z}_1,\ldots,\mathbf{z}_5].
\end{equation}
This supports smooth interpolation across continuous trait space and enables post-hoc trait attribution via attention weights (Fig.~\ref{fig:trait_attention}).

\paragraph{Training objective}
We use two heads (room and co-occurrence) and minimize a weighted sum of room cross-entropy and co-occurrence binary cross-entropy.
We report averages over multiple random seeds.

\subsection{Dual-Layer Human Validation Protocol}
\label{subsec:dual_layer_protocol}

PerSim validates (i) the \emph{behavioral plausibility} of synthesized trajectories and (ii) the \emph{downstream usefulness} of the resulting priors using a unified two-layer protocol (\emph{Round~2}) administered to the same Prolific cohort ($N=200$; anchors are collected in \emph{Round~1}, Sec.~\ref{subsec:human_anchors}), with different stimuli in each layer.

\paragraph{Layer 1: transition plausibility}
Participants rate synthetic transition cards on a 5-point Likert scale (1: very implausible; 5: very plausible).
Each participant evaluates 5 real cards plus one catch trial.
Real cards are stratified to cover all 16 orthogonal trait configurations and multiple phases of the 14-day simulation.
We compare the mean plausibility score to a pre-registered acceptance threshold using a one-sample $t$-test.

\paragraph{Layer 2: predictive utility and rigidity gradient}
Participants perform blinded A/B comparisons between trait-conditioned outputs and the population-frequency baseline, with a ``tie'' option.
Our headline test is a rigidity-modulated gradient: preference for trait-conditioned priors increases as rigidity decreases.
We test this effect via a $\chi^2$ test of independence across rigidity strata (Type A/B/C; Table~\ref{tab:object_rigidity}).
This pattern motivates the rigidity gate in Fig.~\ref{fig:hybrid_policy} and is consistent with rigidity-stratified efficiency trends in downstream digital-twin search (Sec.~\ref{sec:dt_search}).

\paragraph{Gating insight}
Our hypothesis is not personalization dominates universally, but that \emph{utility of trait-conditioned priors increases as rigidity decreases}, yielding a deployable decision rule: use population priors for anchor objects and trait-conditioned priors for behaviorally variable objects.


\begin{table}[b]
\centering
\small
\setlength{\tabcolsep}{4pt}
\caption{Core PerSim dataset statistics after offline validation.}
\label{tab:data_summary}
\resizebox{\columnwidth}{!}{%
\begin{tabular}{@{}l r @{\hskip 8pt} l r@{}}
\toprule
\textbf{Component} & \textbf{Value} & \textbf{Metric} & \textbf{Value} \\
\midrule
Scenes / Personas & 5 / 16 & Layout pass rate & 95.7\% \\
Pers.-layout configs & 80 & Traj.\ pass rate & 92.5\% \\
Traj.\ duration & 14 days & Med./75th dist. & 1.66/2.59\,m \\
Total valid actions & 27{,}976 & Co-occ.\ cov.\ ($\delta$=2.5\,m) & 72.6\% \\
Movable categories & 197 & & \\
\bottomrule
\end{tabular}%
}
\end{table}

\section{Experiments}
\label{sec:experiments}

We evaluate PerSim as a robotics method for \emph{when-to-personalize} household object search.
Our evaluations form a closed loop: (i) validate that synthesized multi-day transitions are behaviorally plausible (Human Layer~1), (ii) test whether trait conditioning is \emph{selectively} useful via a rigidity-modulated gradient (Human Layer~2 and offline prediction), and (iii) show that this selectivity translates into lower end-to-end search effort under a deployable rigidity-gated policy in a home digital twin (ERV/CP@5/ESC).
Throughout, we report \textbf{rigidity-stratified} results (Type~A/B/C) to align human preference, prediction metrics, and downstream search efficiency under the same gating rationale.

\begin{figure*}[t]
  \centering
  \includegraphics[width=0.95\textwidth]{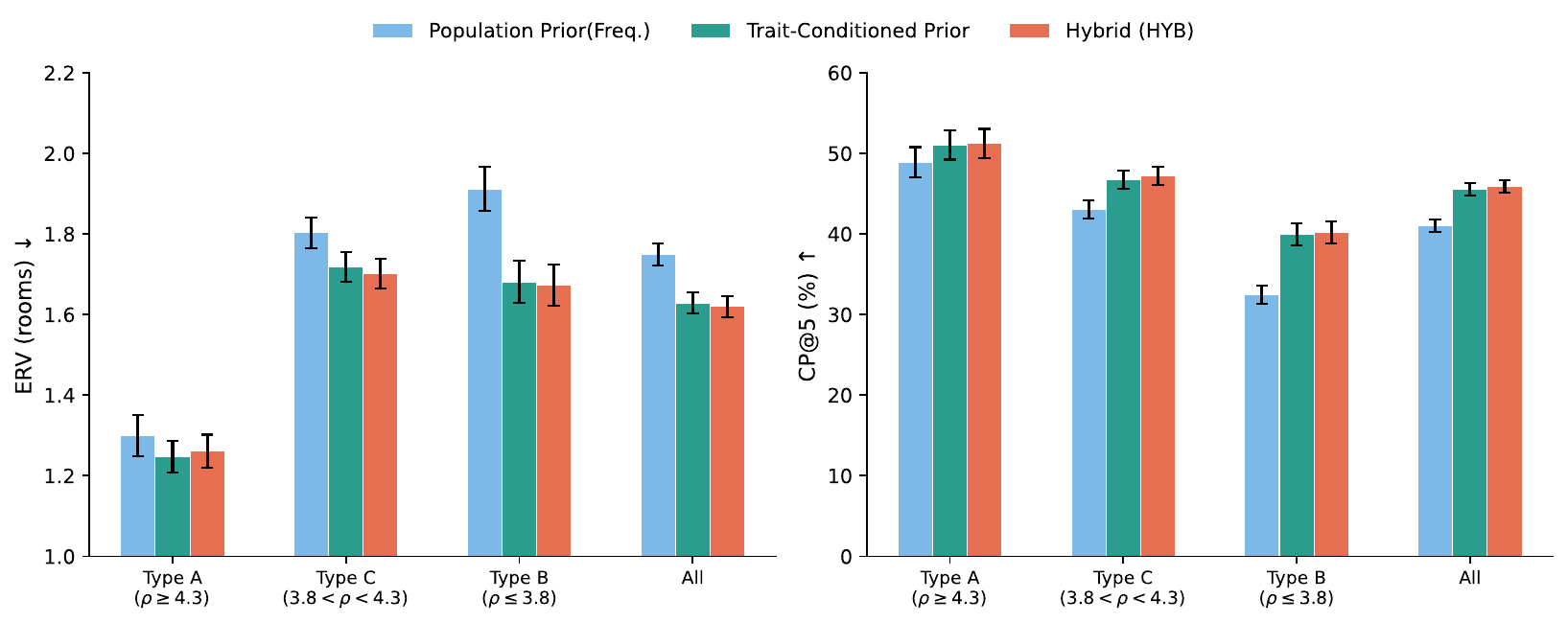}
  \caption{\textbf{Digital-twin two-stage object search with rigidity stratification.}
\textbf{(Left) Room search:} Expected rooms visited (ERV; lower is better) when searching using predicted room priors.
\textbf{(Right) Local search:} Cue Precision@5 (CP@5; higher is better) of predicted within-room co-occurrence cues, evaluated against simulator-derived neighbor sets within radius $\delta$ (Sec.~\ref{subsec:relation_cooccurrence}).
Bars show mean $\pm$ SE over test queries, reported by rigidity type (Table~\ref{tab:object_rigidity}), comparing the population prior (Freq.), trait-conditioned prior, and the rigidity-gated hybrid (HYB).}
\label{fig:dt_search}
\end{figure*}

\subsection{Evaluation Setup and Data Splits}
\label{subsec:data_summary}

\paragraph{Simulated dataset}
PerSim produces 27{,}976 validated object-level actions across 1{,}120 simulated days (80 layout--trait instances) in OmniGibson.
Core statistics are summarized in Table~\ref{tab:data_summary}.
We split actions by \emph{layout--trait instances} to prevent leakage across correlated multi-day trajectories.
Unless otherwise stated, we report mean $\pm$ std over 5 random seeds.

\paragraph{Query objects and rigidity strata}
All evaluations focus on a controlled query set of 15 common, easily misplaced household objects (Table~\ref{tab:object_rigidity}).
Objects are stratified into Type~A/C/B by survey-measured mean rigidity $\rho(o)$ relative to $(\tau_L,\tau_H)$, fixed before any A/B or downstream evaluation---so strata are independent of the outcomes they analyze.

\subsection{Digital-Twin Downstream Search Evaluation}
\label{sec:dt_search}

We translate predicted priors into \emph{robotic search behavior} using the two-stage policy in Sec.~\ref{subsec:dt_policy} (OmniGibson).
Given query $(o,s,\mathbf{t})$, each method outputs (i) a room distribution $\hat{p}(r\mid s,o,\mathbf{t})$ over 8 room types and (ii) within-room cue scores over movable objects.

\paragraph{Metrics: ERV, CP@5, and end-to-end ESC}
We evaluate downstream efficiency using \textbf{ERV} for room ordering (lower is better) and \textbf{CP@5} for within-room cue ranking (higher is better).
Stage~2 is conditioned on the ground-truth room $r^\star$ to isolate local cue-ranking quality.
We report \textbf{Expected Search Cost (ESC)} as a unified effort proxy combining the two stages:
\begin{equation}
\mathrm{ESC}=\mathrm{ERV}+\lambda\cdot k\cdot (1-\mathrm{CP@k}),
\end{equation}
with $\lambda=0.5$ and $k=5$ fixed across methods.
ESC is a policy-aligned cost proxy for comparing methods under a consistent protocol (it does not model physical execution time).

\paragraph{Methods compared}
We compare: \textbf{Population Prior (Freq.)}, \textbf{Trait-Conditioned Prior}, and \textbf{Hybrid (HYB)} (rigidity-gated mixture; Sec.~\ref{subsec:dt_policy}).
Population Prior (Freq.) is deterministic (no learned parameters or seed dependence), so no variance is reported.
HYB is not separately trained; it deterministically mixes the Population Prior (Freq.) and Trait-Conditioned Prior using the rigidity gate.
Uniform/Random are included as sanity-check references.

\paragraph{Results: rigidity-stratified efficiency and end-to-end gains}
Fig.~\ref{fig:dt_search} reports ERV and CP@5 with rigidity stratification.
Across metrics, improvements concentrate on low/medium-rigidity objects (Type~B/C), while gains on anchors (Type~A) are marginal---consistent with the claim that trait conditioning is \emph{conditionally} useful.
Table~\ref{tab:esc} summarizes ESC improvements over the Population Prior (Freq.), showing that HYB achieves the strongest overall end-to-end reduction, indicating that gating converts the rigidity-dependent signal into a more stable search policy.
For within-room cueing, trait conditioning can still help even for anchor-like objects (e.g., via surrounding clutter or container preferences). Since ESC combines ERV with a cue-check penalty derived from CP@5, we report both ERV and CP@5 (and their unified proxy ESC in Table~\ref{tab:esc}) rather than room priors alone.

\begin{table}[t]
\centering
\small
\caption{Expected Search Cost (ESC) improvement over Population Prior (Freq.). $\mathrm{ESC}=\mathrm{ERV}+\lambda\cdot k\cdot(1-\mathrm{CP@k})$, with $\lambda=0.5$, $k=5$. Negative is better.}
\label{tab:esc}
\resizebox{\columnwidth}{!}{%
\begin{tabular}{lcccc}
\toprule
Method & Type A  & Type C  & Type B  & All \\
\midrule
Trait-Conditioned Prior & $-4.1\%$ & $-5.5\%$ & $-11.6\%$ & $-7.3\%$ \\
Hybrid (HYB)            & $-3.7\%$ & $-6.4\%$ & $-12.0\%$ & $-7.8\%$ \\
\bottomrule
\end{tabular}%
}
\end{table}

\begin{table}[b]
\centering
\small
\setlength{\tabcolsep}{3pt}
\caption{Prediction performance (mean $\pm$ std, 5 seeds).}
\label{tab:ablation}
\resizebox{\columnwidth}{!}{%
\begin{tabular}{@{}lcccc@{}}
\toprule
\textbf{Model} & \textbf{Room@1} & \textbf{Room@2} & \textbf{P@5} & \textbf{R@5} \\
\midrule
Baseline (Random) & 12.9 {\scriptsize$\pm$.5} & 25.3 {\scriptsize$\pm$.6} & 4.3 {\scriptsize$\pm$.2} & 2.5 {\scriptsize$\pm$.2} \\
\textbf{Population (Freq.)} 
& 57.6 & 80.7 & 41.0 & 25.5 \\
\textbf{Trait-Conditioned (Ours)} 
& \textbf{66.4} {\scriptsize$\pm$.3} 
& \textbf{85.6} {\scriptsize$\pm$.2} 
& \textbf{47.8} {\scriptsize$\pm$.3} 
& \textbf{28.5} {\scriptsize$\pm$.3} \\
\bottomrule
\end{tabular}%
}
\end{table}

\begin{table}[t]
\centering
\small
\caption{Rigidity-stratified Room@1 accuracy (macro-averaged within each type). $\Delta$: Freq.$\to$Ours improvement.}
\label{tab:rigidity_stratified}
\setlength{\tabcolsep}{4pt}
\begin{tabular}{lcccc}
\toprule
& Freq. & Trait. (Ours) & $\Delta$ \\
\midrule
Type A (Anchor)  & 78.4\% & 78.0\% & $-$0.4 pp \\
Type C (Moderate)  & 52.1\% & 54.3\% & +2.0 pp \\
Type B (Sensitive)  & 47.8\% & 61.0\% & +13.2 pp \\
\bottomrule
\end{tabular}
\vspace{-2pt}
\end{table}

\begin{figure*}[t]
    \centering
    \begin{minipage}[t]{0.46\textwidth}
        \centering
        \includegraphics[width=\linewidth]{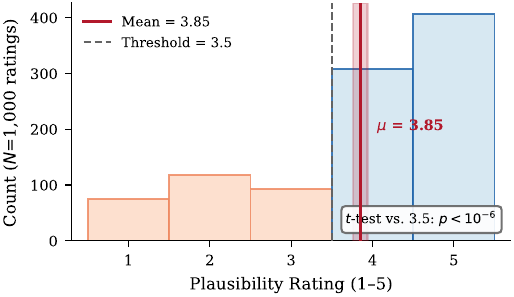}
    \end{minipage}\hfill
    \begin{minipage}[t]{0.46\textwidth}
        \centering
        \includegraphics[width=\linewidth]{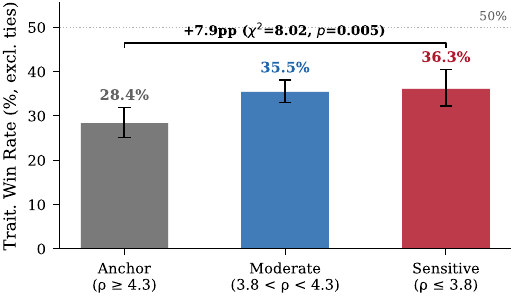}
    \end{minipage}
    \caption{\textbf{Human validation of the PerSim pipeline.}
\textbf{(Left) Layer-1: Synthetic transition plausibility.}
Across $N{=}1{,}000$ ratings, sampled transitions achieve mean plausibility $3.85/5$ (95\% CI [3.773, 3.931]), exceeding the acceptance threshold 3.5 ($p<10^{-6}$), supporting the use of synthetic dynamics as training signals.
\textbf{(Right) Layer-2: Rigidity-modulated utility of personalization.}
In a blinded A/B comparison (ties 15.3\%), trait conditioning is preferred more for low-rigidity objects than for high-rigidity anchors (36.3\% vs 28.4\% win rate excl.\ ties; +7.9pp; $\chi^2{=}8.02$, $p{=}0.005$). Error bars: 95\% binomial CIs.}
    \label{fig:human_validation}
\end{figure*}

\subsection{Offline Prediction Quality on Simulated Actions}
\label{subsec:offline_results}

We evaluate prediction quality on simulated actions to isolate model capacity independent of rollout.
We report room prediction accuracy (Room@1/Room@2) and cue prediction quality (P@5/R@5) (Table~\ref{tab:ablation}).
Our \textbf{primary reporting} uses rigidity-stratified macro-averages (Type~A/B/C), consistent with the hybrid policy rationale; we include overall aggregates for completeness.
Per-object results (omitted for space) reveal that trait conditioning helps most on a few variable-placement items and may introduce noise for some moderate items with limited samples; this heterogeneity directly motivates the rigidity-gated hybrid policy that stabilizes end-to-end performance (Table~\ref{tab:esc}).

\paragraph{Main results}
Table~\ref{tab:rigidity_stratified} reports rigidity-stratified offline performance for the Population Prior (Freq.) and Trait-Conditioned Prior.
Trait conditioning yields the largest gains on sensitive objects (Type~B), smaller or mixed gains on moderate objects (Type~C), and limited gains on anchors (Type~A).
This rigidity-modulated pattern mirrors downstream digital-twin efficiency (Sec.~\ref{sec:dt_search}), while the deployable HYB policy further converts this heterogeneity into improved end-to-end search cost (Table~\ref{tab:esc}).

\subsection{Dual-Layer Human Validation}
\label{subsec:human_validation_results}

We validate PerSim (Fig.~\ref{fig:human_validation}) using a unified protocol administered to the same Prolific cohort ($N=200$).

\paragraph{Layer 1: transition plausibility}
Synthetic transitions are judged behaviorally plausible with a mean rating of \textbf{3.85/5} (95\% CI [3.773, 3.931], $N{=}1000$),
significantly exceeding the 3.5 acceptance threshold (one-sample $t$-test, $p<10^{-6}$).
This establishes that calibrated generation produces plausible household dynamics under our protocol.

\paragraph{Layer 2: predictive utility and rigidity gradient}
In blinded A/B comparisons, we report win rates excluding ties (tie rate 15.3\%).
The overall win rate is \emph{not} the headline; instead, we test the rigidity-modulated gradient.
Preference for the Trait-Conditioned Prior increases as rigidity decreases: win rate rises from 28.4\% on anchor objects (Type~A) to \textbf{36.3\%} on sensitive objects (Type~B), a gradient of \textbf{+7.9pp} ($\chi^2=8.02$, $p=0.005$).
This human-facing gradient is consistent with rigidity-stratified downstream efficiency (Sec.~\ref{sec:dt_search}) and directly motivates the rigidity-gated HYB policy. Win rates stay below 50\% even for sensitive objects because the population prior captures genuinely shared regularities; our claim concerns the \emph{marginal} gain of personalization as rigidity falls, not absolute dominance---which is exactly what the gate exploits.

\begin{table}[t]
\centering
\small
\setlength{\tabcolsep}{4pt}
\renewcommand{\arraystretch}{0.92}
\caption{Generalization to unseen continuous traits ($N{=}200$; offline).}
\label{tab:gen}
\begin{tabular}{@{}lcc@{}}
\toprule
\textbf{Conditioning} & \textbf{Hit@1} & \textbf{P@5} \\
\midrule
Matched (L16)                & 60.2 & 23.1 \\
\textbf{Cont.\ (Actual traits)} & \textbf{61.1} & \textbf{23.2} \\
\midrule
$\Delta$     & \textbf{+0.9} ($p{=}0.035$) & +0.1 (ns) \\
\bottomrule
\multicolumn{3}{@{}l}{\scriptsize Hit@1: Room accuracy (\%). P@5: Co-occ.\ precision (\%). ns: not significant.}
\end{tabular}
\end{table}

\subsection{Generalization and Interpretability}
\label{subsec:generalization}

\paragraph{Continuous trait interpolation}
To test generalization beyond the 16 discrete training configurations, we compare conditioning on participants' \emph{actual} continuous trait vectors (Cont.) versus nearest discrete configuration matching (Matched), evaluated offline against participants' self-reported ground truth.
As shown in Table~\ref{tab:gen}, conditioning on continuous traits yields a small but significant improvement ($p=0.035$), supporting interpolation in five-dimensional trait space.

\paragraph{Interpretability}
Cross-attention weights provide inspectable (heuristic) trait attributions for model decisions.
We visualize object-specific trait dependence patterns (Fig.~\ref{fig:trait_attention} and Fig.~\ref{fig:room_priors_regimes}) and provide qualitative examples as post-hoc rationales for why the robot prioritizes specific locations under a given resident profile.

\section{Discussion}
\label{sec:discussion}
PerSim addresses a practical question: \emph{when should a robot personalize its priors, and when default to population norms?}
Across offline prediction, digital-twin search, and dual-layer human evaluation, we find personalization is \emph{item-dependent} rather than universally beneficial---trait conditioning helps behaviorally variable objects, while population priors suffice for anchor-like items with near-universal placement.
This motivates \textbf{selective hybrid priors} that gate on an item-level sensitivity estimate such as rigidity, consistent with the observed preference gradient (+7.9pp, $p=0.005$).

\section{Limitations}
\label{sec:limitations}

PerSim is evaluated in a digital twin and via human judgments; real homes include additional factors (e.g., clutter, shared spaces, and idiosyncratic storage rules) that may shift placement distributions.
Our resident profile is restricted to Big Five traits; behavior can also depend on culture, household composition, routines, and accessibility constraints.
We study \emph{search priors} and their induced search efficiency, not full embodied execution; on a physical robot, PerSim serves as a plug-in prior generator that maps a query and semantic map (optionally a resident profile) to room-ranking and cue priors used by standard navigation/search stacks.
Rigidity is self-reported and evaluated offline; longitudinal in-home studies would better capture adaptation and habit formation.
A stronger trait-agnostic baseline (e.g., a VLM/LLM semantic room prior) and learning the gate end-to-end are promising directions.

\section{Conclusion}
\label{sec:conclusion}

We presented PerSim, a human-calibrated framework that learns trait-conditioned spatial priors for household robot object search.
Results show that trait conditioning is \emph{not} universally helpful: benefits concentrate on behaviorally variable objects and follow a rigidity-modulated gradient ($p=0.005$), motivating a selective hybrid strategy that defaults to population priors for anchor items.
Concretely, we apply trait conditioning when an object's rigidity score falls below a threshold and otherwise rely on population priors, with a smooth interpolation in the intermediate regime.
We also find evidence of generalization to unseen continuous trait vectors ($p=0.035$), supporting interpolation in a five-dimensional resident-profile space.
In the accompanying video, we provide qualitative rollouts illustrating when the rigidity gate defaults to population norms versus when personalization meaningfully shifts search.

\section*{Acknowledgment}
This study was supported by the NVAITC and UFIT, University of Florida, and approved by the University of Florida IRB (Protocol \#:~ET00049546). In compliance with IEEE policy on AI-generated content, we disclose that the synthetic placement transitions used in this work were generated using Gemini~2.5~Flash (Google Cloud Vertex AI), with human-anchored calibration and validation as described in Sec.~\ref{sec:method}.






\balance
\bibliographystyle{IEEEtran}
\bibliography{refs}

@article{zhang2019efficient,
  title={Efficient dynamic object search in home environment by mobile robot: A priori knowledge-based approach},
  author={Zhang, Ying and Tian, Guohui and Lu, Jiaxing and Zhang, Mengyang and Zhang, Senyan},
  journal={IEEE Transactions on Vehicular Technology},
  volume={68},
  number={10},
  pages={9466--9477},
  year={2019},
  publisher={IEEE}
}

@inproceedings{pangercic2012semantic,
  title={Semantic object maps for robotic housework--representation, acquisition and use},
  author={Pangercic, Dejan and Pitzer, Benjamin and Tenorth, Moritz and Beetz, Michael},
  booktitle={2012 IEEE/RSJ International Conference on Intelligent Robots and Systems},
  pages={4644--4651},
  year={2012},
  organization={IEEE}
}

@article{2024seek,
  title={SEEK: Semantic reasoning for object goal navigation in real world inspection tasks},
  author={Ginting, Muhammad Fadhil and Kim, Sung-Kyun and Fan, David D and Palieri, Matteo and Kochenderfer, Mykel J and Agha-Mohammadi, Ali-akbar},
  journal={arXiv preprint arXiv:2405.09822},
  year={2024}
}

@inproceedings{sarch2022tidee,
  title={TIDEE: Tidying up novel rooms using visuo-semantic commonsense priors},
  author={Sarch, Gabriel and Fang, Zhaoyuan and Harley, Adam W and Schydlo, Paul and Tarr, Michael J and Gupta, Saurabh and Fragkiadaki, Katerina},
  booktitle={European Conference on Computer Vision},
  pages={480--496},
  year={2022},
  organization={Springer}
}

@inproceedings{kant2022housekeep,
  title={Housekeep: Tidying virtual households using commonsense reasoning},
  author={Kant, Yash and Ramachandran, Arun and Yenamandra, Sriram and Gilitschenski, Igor and Batra, Dhruv and Szot, Andrew and Agrawal, Harsh},
  booktitle={European Conference on Computer Vision},
  pages={355--373},
  year={2022},
  organization={Springer}
}

@inproceedings{chaplot2020object,
  title={Object goal navigation using goal-oriented semantic exploration},
  author={Chaplot, Devendra Singh and Gandhi, Dhiraj Prakashchand and Gupta, Abhinav and Salakhutdinov, Ruslan},
  booktitle={Advances in Neural Information Processing Systems},
  volume={33},
  pages={4247--4258},
  year={2020}
}

@article{wu2023tidybot,
  title={TidyBot: Personalized robot assistance with large language models},
  author={Wu, Jimmy and Antonova, Rika and Kan, Adam and Lepert, Marion and Zeng, Andy and Song, Shuran and Bohg, Jeannette and Rusinkiewicz, Szymon and Funkhouser, Thomas},
  journal={Autonomous Robots},
  volume={47},
  number={8},
  pages={1087--1102},
  year={2023},
  publisher={Springer}
}

@article{abdo2016organizing,
  title={Organizing objects by predicting user preferences through collaborative filtering},
  author={Abdo, Nichola and Stachniss, Cyrill and Spinello, Luciano and Burgard, Wolfram},
  journal={The International Journal of Robotics Research},
  volume={35},
  number={13},
  pages={1587--1608},
  year={2016},
  publisher={SAGE}
}

@article{gosling2002room,
  title={A room with a cue: Personality judgments based on offices and bedrooms},
  author={Gosling, Samuel D and Ko, Sei Jin and Mannarelli, Thomas and Morris, Margaret E},
  journal={Journal of Personality and Social Psychology},
  volume={82},
  number={3},
  pages={379--398},
  year={2002},
  publisher={American Psychological Association}
}

@article{john1999bigfive,
  title={The Big-Five trait taxonomy: History, measurement, and theoretical perspectives},
  author={John, Oliver P and Srivastava, Sanjay},
  journal={Handbook of Personality: Theory and Research},
  volume={2},
  pages={102--138},
  year={1999},
  publisher={Guilford Press}
}

@article{rammstedt2007bfi10,
  title={Measuring personality in one minute or less: A 10-item short version of the Big Five Inventory in English and German},
  author={Rammstedt, Beatrice and John, Oliver P},
  journal={Journal of Research in Personality},
  volume={41},
  number={1},
  pages={203--212},
  year={2007},
  publisher={Elsevier}
}

@inproceedings{li2024behavior,
  title={Behavior-1k: A benchmark for embodied ai with 1,000 everyday activities and realistic simulation},
  author={Li, Chengshu and Zhang, Ruohan and Wong, Josiah and Gokmen, Cem and Srivastava, Sanjana and Mart{\'\i}n-Mart{\'\i}n, Roberto and Wang, Chen and Levine, Gabrael and Lingelbach, Michael and Sun, Jiankai and others},
  booktitle={Conference on Robot Learning},
  pages={80--93},
  year={2023},
  organization={PMLR}
}

@inproceedings{park2023generative,
  title={Generative agents: Interactive simulacra of human behavior},
  author={Park, Joon Sung and O'Brien, Joseph and Cai, Carrie Jun and Morris, Meredith Ringel and Liang, Percy and Bernstein, Michael S},
  booktitle={Proceedings of the 36th Annual ACM Symposium on User Interface Software and Technology},
  pages={1--22},
  year={2023}
}

@article{ahn2022saycan,
  title={Do as I can, not as I say: Grounding language in robotic affordances},
  author={Ahn, Michael and Brohan, Anthony and Brown, Noah and Chebotar, Yevgen and Cortes, Omar and David, Byron and Finn, Chelsea and Fu, Chuyuan and Gopalakrishnan, Keerthana and Hausman, Karol and others},
  journal={arXiv preprint arXiv:2204.01691},
  year={2022}
}

@inproceedings{wang2023voyager,
  title={Voyager: An open-ended embodied agent with large language models},
  author={Wang, Guanzhi and Xie, Yuqi and Jiang, Yunfan and Mandlekar, Ajay and Xiao, Chaowei and Zhu, Yuke and Fan, Linxi and Anandkumar, Anima},
  booktitle={Advances in Neural Information Processing Systems},
  volume={36},
  year={2023}
}

@inproceedings{huang2023inner,
  title={Inner monologue: Embodied reasoning through planning with language models},
  author={Huang, Wenlong and Xia, Fei and Xiao, Ted and Chan, Harris and Liang, Jacky and Florence, Pete and Zeng, Andy and Tompson, Jonathan and Mordatch, Igor and Chebotar, Yevgen and others},
  booktitle={Conference on Robot Learning},
  pages={1769--1782},
  year={2023},
  organization={PMLR}
}

@inproceedings{unal1990taguchi,
  title={Taguchi approach to design optimization for quality and cost: An overview},
  author={Unal, Resit and Dean, Edwin B},
  booktitle={Proceedings of the Annual Conference of the International Society of Parametric Analysts},
  pages={28--32},
  year={1991}
}

@article{kacker1991taguchi,
  title={Off-line quality control, parameter design, and the Taguchi method},
  author={Kackar, Raghu N},
  journal={Journal of Quality Technology},
  volume={17},
  number={4},
  pages={176--188},
  year={1985}
}

@inproceedings{vaswani2017attention,
  title={Attention is all you need},
  author={Vaswani, Ashish and Shazeer, Noam and Parmar, Niki and Uszkoreit, Jakob and Jones, Llion and Gomez, Aidan N and Kaiser, {\L}ukasz and Polosukhin, Illia},
  booktitle={Advances in Neural Information Processing Systems},
  volume={30},
  year={2017}
}

\end{document}